# A funny companion: Distinct neural responses to AI- versus human-attributed humor

Xiaohui Rao[1], Hanlin Wu[1], Zhenguang G. Cai[1, 2,*]

[1] Department of Linguistics and Modern Languages [2] Brain and Mind Institute

The Chinese University of Hong Kong Hong Kong SAR

## Data availability

We have uploaded all the stimuli, data and analysis scripts onto Open Science Framework (https://osf.io/8796a/?view_only=0605879d5b91482f8539e79a204876ce).

## Funding

This work was supported by General Research Fund (14601525), University Grants Committee, Hong Kong, Do people treat personalized large language models as humans in communication?

## Declaration of competing interest

The authors declare that they have no known competing financial interests or personal relationships that could have appeared to influence the work reported in this paper.

## Author notes

Correspondence should be addressed to Zhenguang Cai, Department of Linguistics and Modern Languages, Leung Kau Kui Building, The Chinese University of Hong Kong, Shatin, N.T., Hong Kong; zhenguangcai@cuhk.edu.hk.

## ABSTRACT

As artificial intelligence (AI) companions become capable of human-like communication, including telling jokes, understanding how people cognitively and affectively respond to AI-attributed humor becomes increasingly important. This study used electroencephalography (EEG) to compare how people process culturally embedded puns versus controls attributed to either an AI agent or a human. In an interaction task, participants guessed punchlines based on joke setups before they were disclosed by the interlocutor. Behavioral analysis revealed that participants rated AI- and human-attributed humor as comparably funny. However, neurophysiological data showed that AI-attributed humor elicited a reduced and sustained N400 effect compared to human-attributed humor, suggesting reduced cognitive effort in semantic conflict detection and resolution, or attenuated feedback-related processing. This was also accompanied by a larger late positive potential (LPP), reflecting intensified late-stage affective and evaluative processing. This likely reflects a process where the brain reconciles AI's delivery of culturally embedded puns with prior low expectations, leading to intensified affective engagement and interlocutor model updating. Individual differences in social perceptions further influenced neural responses. Higher perceived AI sincerity and trustworthiness were associated with a globally reduced N400, facilitating semantic integration for both humorous and non-humorous AI language. Additionally, increased AI trustworthiness predicted an enhanced LPP, indicating a more intensified updating of the interlocutor model during humor comprehension. These findings indicate that the brain's neural sensitivity to AI-attributed humor bypasses biases like algorithm aversion. This highlights the brain's adaptation to humor from a novel source and underscores humor's potential for fostering genuine engagement in human-AI social interaction.

## 1. INTRODUCTION

The rise of Large Language Models (LLMs) has marked a leap in artificial intelligence (AI), creating social agents capable of human-like language communication (e.g., Cai et al., 2024). These models are no longer just tools for text generation; they are increasingly integrated into social contexts, serving as psychological chatbots for mental health support (Casu et al., 2024), virtual tutors in educational settings (García-Méndez et al., 2024), reading companions for children (Liu et al., 2022), and AI friends (Maples et al., 2024). This integration raises questions about the nature of human-AI social interaction and the cognitive and affective processes at play when humans engage with content traditionally considered the exclusive domain of human expressions.

One of the most complicated forms of human expressions is humor. Its comprehension and appreciation are not just about understanding words, requiring the ability to detect and resolve incongruity, social cognition and the interplay of various individual factors (Canal et al., 2019). Traditionally viewed as a unique human trait and a social glue that fosters cooperation and reduces social distance (Sherman, 1988), humor's landscape is evolving. Notably, AI companions are now capable of generating humor, such as puns (Chen et al., 2024), with some AI-generated jokes even indistinguishable from those created by humans (Gorenz & Schwarz, 2024; Joshi, 2025). However, the social functions of humor may hinge not only on its quality but also on the perceived source and intent (Martin & Ford, 2018). This leads to a question: when humans encounter high-quality AI-attributed humor, does the brain's processing differ based on whether the source is perceived as artificial?

A challenge for AI companions is *algorithm aversion*, defined as the human tendency to prefer human over algorithmic judgment (Castelo et al., 2019; Mahmud et al., 2022). This aversion frequently occurs in social interaction, as it is a domain requiring human traits like personal taste, intuition, and experience (Bower & Steyvers, 2021). The findings from a series of nine studies by Rubin et al. (2025) illustrate this challenge: responses believed to be AI-generated were perceived to be less empathetic and supportive, and more affectively negative, than those believed to be human-generated. This bias is largely driven by knowing the source of the response. For example, Yin et al. (2024) had participants describe a complex personal situation and receive support. When participants did not know who wrote the response, they actually felt “heard” more often by AI than by humans. However, once the source was labeled, the effect reversed, and participants preferred the human responses. While AI can produce highly effective expressions of empathy, these responses are dismissed as simulated once the source is known. AI models are perceived as lacking a genuine cognitive, affective, or motivational state (Chen et al., 2023; Rao et al., 2025; Rubin et al., 2024). They do not share our pain or joy, and their output requires no effort on their part (Li et al., 2026; Perry, 2023; Shteynberg et al., 2024). This lack of authentic caring, a key component of human empathy, may be why disclosing AI involvement diminishes the perceived social and emotional benefits of an interaction (Rubin et al., 2025).

Interestingly, while human-AI interaction requiring empathy suffer from a critical “caring” deficit, this bias is not absolute in the domain of humor. Bower and Steyvers (2021) found that, when asked to rate jokes and guess their likeliest source, participants rated jokes they thought were human-generated as the funniest while those thought to be AI-generated ranked the lowest; they tended to believe a human-generated joke to be AI-generated when the joke was low in quality. However, when joke sources were disclosed, this aversion disappeared: identical jokes received similar funniness ratings regardless of whether they were labeled as human- or AI-created. This indicates that even if there are aversions toward AI in this domain, they are easily overcome when presented with counterevidence such as well-received jokes framed as AI-created (Bower & Steyvers, 2021). Furthermore, people’s appreciation for AI-generated humor increases with repeated exposure (Joshi, 2025), indicating these attitudes are malleable. One possible reason for such malleability is that biases against AI are easily countered by evidence of its competence (Bower & Steyvers, 2021). Additionally, humor appreciation may be driven more by the cognitive surprise of the joke’s resolution (Suls, 1972) than by an empathetic bond with the source. Consequently, humor may be a domain where AI can achieve social success, offering a pathway to mitigate algorithm aversion and reduce social distance between AI companions and humans.

Early research found that machine joke-telling differs from human joke-telling on two key dimensions: joke quality and perceived sociability. Binsted and Ritchie (1994) found that jokes produced by AI of the time were rated low in humorous quality by school children. Concurrently, Morkes et al. (1999) reported that people felt less sociable and experienced less mirth in response to humor when interacting with a computer compared to communicating with another human via a computer, suggesting that low machine sociability hindered effective joke-telling. However, as AI has evolved into a more advanced social agent, a shift accelerated by the emergence of LLMs, recent studies indicate that the gap between human and machine joke-telling is narrowing. Avetisyan et al. (2023) demonstrated AI's capacity to generate humorous and novel content. This is further supported by Gorenz and Schwarz (2024), who found that jokes generated by ChatGPT 3.5 were rated as equally humorous as (and sometimes even more humorous than) human-generated jokes in a masked evaluation. Similarly, Joshi (2025) reported that participants struggled to differentiate between human- and AI-generated jokes.

Despite the development of LLM-powered AI in generating humorous content, how people cognitively and affectively comprehend AI-attributed humor compared to human-attributed one remains unexplored. Suls's (1972) two-stage model of humor comprehension, supported by electroencephalography (EEG) studies on human-human interaction (e.g., Canal et al., 2019), provides a useful framework. The first stage is related to incongruity detection. Take for instance the pun *A bicycle cannot stand on its own because it is two-tired* (Zheng & Wang, 2023). Comprehenders simultaneously activates the pun's literal meaning, which is congruent with the bicycle's two wheels, and its homophonic counterpart, "too tired". This second meaning, though literally incongruous for an inanimate object, humorously applies a human state to the bicycle. The second stage involves semantic conflict resolution, which integrates cognitive, affective, and evaluative components. The cognitive process involves incongruity resolution and integration (Ku et al., 2017; Shibata et al., 2017). This is followed by affective and evaluative processes whereby comprehenders experience mirth and form a subjective appraisal of the joke's delivery (Chang et al., 2018; Du et al., 2013; Ku et al., 2017).

Building on this framework, the present study targets a specific type of humor: homophonic/character-substitution puns in Chinese (Feng et al., 2014; Li et al., 2025). These puns are operationally defined by the necessary features of phonological similarity and orthographic substitution and often include the optional feature of relying on idiomatic expressions and/or internet slang (see the Methods section for examples). Comprehending this specific humor type requires a sequence of cognitive processes. First, comprehenders need to detect phonological/orthographic ambiguity and activate competing lexical candidates; this process is usually indexed by the N400 component, a negative deflection peaking around 400 ms post-stimulus. The N400 is sensitive to predictability and semantic congruence (Kutas & Federmeier, 2011), reflecting the brain's initial response to the semantic incongruity within the pun (Coulson & Kutas, 2001; Ku et al., 2017). Second, comprehenders need to retrieve the relevant idiom or slang form to resolve the intended expression, and reinterpret the setup to integrate competing interpretations; such semantic conflict resolution is often associated with the P600, a positive-going wave reflecting late-stage reanalysis (Mayerhofer & Schacht, 2015). While this stage is often indexed by the P600, some studies observe a sustained N400. The prolonged negativity is thought to reflect extended semantic conflict processing, where the brain continues to recruit cognitive resources to integrate competing interpretations (Coulson & Kutas, 2001; Mayerhofer & Schacht, 2015). Finally, as comprehenders successfully integrate competing interpretations (e.g., resolving the substituted character in a Chinese pun), the resulting affective and evaluative processing during humor appreciation is typically reflected by the late positive potential (LPP; Chang et al., 2018; Du et al., 2013; Ku et al., 2017).

Importantly, humor comprehension is likely influenced by expectations about the joke-teller. Johnson and Mistry (2013) found that, when a joke was attributed to a named comedian, prior exposure to that individual primed an expectation of humor, which enhanced subsequent humor ratings compared to jokes attributed to non-comedians. This aligns with the theory of interlocutor modelling, where people construct a model of their conversation partner to adjust their language production (Cai et al., 2021; Shen & Wang, 2023; Zhang & Cai, 2026) and comprehension (Cai, 2022; Cai et al., 2017; Wu & Cai, 2026; Wu et al., 2024). Interlocutor modelling, or the integration of the speaker identity into language comprehension, may be relevant for human-AI interaction. Rao et al. (2025b) demonstrated that humans process AI-attributed semantic and syntactic anomalies differently from those attributed to humans. This difference appears to stem from the perception that AI excels at grammatical competence but is prone to semantic hallucination. Consequently, readers exhibited a reduced N400 effect for semantic anomalies and an enhanced P600 effect for syntactic anomalies in texts believed to be AI-generated. These findings suggest that the human brain engages distinct expectations when processing language produced by AI. Extending this insight, it is plausible that humor comprehension involving an AI agent also recruits unique cognitive pathways. Understanding these neural distinctions is essential for designing AI companions capable of fostering meaningful human connection.

To move beyond subjective funniness ratings, which are susceptible to reporting biases (Silvia et al., 2021; Suls, 1983), our study utilizes EEG to examine the cognitive and affective processes when comprehending AI- versus human-attributed humor. Based on the established two-stage model of humor processing (Suls, 1972) and the theory of interlocutor modelling (Cai et al., 2017; Wu & Cai, 2026; Wu et al., 2024), we hypothesize that the brain's cognitive responses to humor will differ depending whether the joke is attributed to humans or AI. In particular, recent research shows that people's expectations of AI capabilities, such as its propensity for semantic hallucination (Maynez et al., 2020; Rawte et al., 2023) and syntactic proficiency, modulate their language processing (Rao et al., 2025b). Applying these findings to humor, we predict that people are less sensitive to semantic incongruity in AI-attributed humor, given that they may have lower expectations for AI's semantic and logical consistency (Rao et al., 2025b) or for its competence in generating humor with complex culturally embedded puns, leading to a smaller N400 for AI-attributed humor compared to human-attributed humor. We also predict that people exert less cognitive effort during conflict resolution, indexed by a smaller P600 or an attenuated long-lasting N400 in response to AI humor than human humor, as people may not be as engaged in fully processing the joke's intended meaning when they know it comes from a non-sentient AI. We anticipate that the LPP, a marker of late-stage affective and evaluative processing, will be larger for AI-attributed humor, as the brain may reconcile AI's successful delivery of these puns with its prior low expectations of the agent. Finally, we explore whether people's social perceptions of AI, such as perceived trustworthiness (Flanagin & Metzger, 2003) and sincerity (Liu & Sundar, 2018), modulate these neural responses.

## 2. METHODS

### *2.1 Participants*

Sixty-four neurologically healthy, native Mandarin speakers (28 males; mean age 23.0 years, range 18-35) were initially recruited. Seven participants were excluded from data analyses: one for failing the manipulation check and six due to excessive EEG artifacts (see Sections 2.3 and 2.4 for details). This resulted in a final sample of 57 participants (25 males; mean age 22.6 years, range 18-35). All participants gave informed written consent before participation. The study protocol was approved by the relevant Joint Clinical Research Ethics Committee.

### *2.2 Design and materials*

We adopted a 2 (Stimulus: humor vs control) × 2 (Attribution: AI vs human) factorial design. Stimulus was manipulated within both participants and items, while Attribution was manipulated between participants and within items.

Ninety Mandarin homophonic/character-substitution puns were developed for the humor condition. These were either sourced from online sources and books or adapted from prior research (e.g., Feng et al., 2014; Li et al., 2025). Each stimulus followed a question-and-answer format: a setup and a punchline. This design mirrors English homophonic puns where a word in the setup primes the reinterpretation of a fixed idiom; for instance, a setup asking why a knight stays awake triggers the substituted punchline, "to enjoy the night life". Similarly, our stimuli utilized Mandarin phonological similarity and orthographic substitution to trigger incongruity. The setups contained 10-17 characters and punchlines containing exactly four characters. As shown in Table 1, the punchline is a well-known idiom (e.g., 归心似箭, $gui_{1}xin_{1}si_{4}jian_{4}$, meaning "yearning to return home immediately"). In the control condition, the setup leads to this standard idiomatic meaning. However, in the humor condition, the setup introduces a phonological bias; for instance, by mentioning a "turtle" (乌龟, $wu_{1}gui_{1}$), the setup forces the reader to reinterpret the first character of the idiom (归, $gui_{1}$, return) as its homophone (龟, $gui_{1}$, turtle). This transforms the idiom's meaning into a literal, absurd statement: "The heart of a turtle is shaped like an arrow". This design requires comprehenders to detect the phonological/orthographic ambiguity, activate competing lexical candidates, retrieve a relevant idiom or slang form, and reinterpret the setup to integrate the competing interpretations. We excluded aggressive and dark humors from the materials, reducing the elicited negative connotations and emotions. In a pretest, one hundred puns were rated by 20 participants who were not involved in the main experiment on a 5-point Likert scale regarding to what extent they thought a pun was comprehensible and funny (1: not at all; 5: very much). Puns with an average rating below 3 in either comprehensibility (*How comprehensible is the joke?*) or funniness (*How funny is the joke?*) were discarded. We eventually selected 90 puns for the experiment (see Table 1 for example), with an average rating of 4.26 ($SD$ = 0.45) for comprehensibility and an average of 3.69 for funniness ($SD$ = 0.37).

**Table 1**. Examples of the experimental materials.

| **Stimulus** | **Setup sentence** | **Punchline** |
|---|---|---|
| Humor | 乌龟 的 心 是 弓箭 形状 的。<br>$Wu_1gui_1$ *de* $xin_1$ $shi_4$ $gong_1jian_4$ $xing_2zhuang_4$ *de.*<br>The heart of a turtle is shaped like an arrow. | 归心似箭<br>$gui_1xin_1si_4jian_4$<br>yearn to return home immediately |
| Control | 渴望 回家 的 心 非常 强烈。<br>$Ke_3wang_4$ $hui_2jia_1$ *de* $xin_1$ $fei_1chang_2$ $qiang_2lie_4$.<br>The longing for home is incredibly strong. | 归心似箭<br>$gui_1xin_1si_4jian_4$<br>yearn to return home immediately |

*Note:* 归 (return, $gui_1$) is phonetically identical to 龟 (turtle, $gui_1$) . Combined with 心(heart, $xin_1$) and 似箭 (be like an arrow, $si_4jian_4$), the expression yields a homophonic alternative meaning that "The heart of a turtle is shaped like an arrow".

### *2.3 Procedure*

**Questionnaires.** Participants reported their general perception of AI chatbots in terms of trustworthiness and sincerity on a 1 to 7 semantic differential scale (i.e., a type of rating scale that uses bipolar adjectives) based on their previous experience. Firstly, in line with previous studies (e.g., Flanagin & Metzger, 2003; Rheu et al., 2024), three items (dishonest to honest, untrustworthy to trustworthy, unreliable to reliable) were used to measure participants' trustworthiness of AI chatbots. Secondly, three items (insincere to sincere, artificial to genuine, performed to heartfelt) were used to evaluate participants' perception of sincerity of AI chatbots (see also Liu & Sundar, 2018; Rheu et al., 2024). Participants in the AI condition completed these measures before the EEG experiment to capture initial perceptions, while those in the human condition completed them post-experiment to avoid compromising interaction authenticity.

**EEG experiment.** We employed the Wizard-of-Oz method to mitigate the influence of the confounding factors (Kelley, 1984): participants were introduced to believe that they were interacting with an AI chatbot or a human peer, which was, in fact, a procedure via E-Prime 3.0 controlled by the experimenter. Participants, fitted with an EEG cap, were placed in a soundproof booth. They were divided into two groups: the AI condition and the human condition. Those in the AI condition were informed they would read jokes told by an AI companion (chatbot), and, to reinforce this, first interacted with the AI companion (i.e., an in-house interface with GPT-4 API) in Mandarin Chinese for three minutes. Participants in the human group, conversely, were told they would read jokes told by a human peer. As illustrated in Figure 1, a fixation first appeared for 1 second, during which participants were told that their partner was typing; then, a picture of an AI companion or a human appeared, which reinforced participants' impression of their interlocutor. The human interlocutors consisted of both a male and a female to ensure a balanced design. At the same time, a setup sentence "sent by the interlocutor" appeared for 10 seconds, during which participants need to guess a four-character expression based on the joke setup. They were required to press the spacebar as soon as they had formulated a covert guess. After a response or time-out, punchline (a four-character expression in Microsoft YaHei in size 36) would be "disclosed by the interlocutor" and displayed for 3 seconds at the center of the screen. The punchline was the critical word and the main focus of the EEG analyses. Participants read the punchline and then indicated whether their prior guess was correct, incorrect, or if they had failed to make a guess. They then rated to what extent they thought this joke was funny and comprehensible (on two separate ratings) on a 5-point Likert scale.

**Manipulation check for interlocutor.** To check whether participants believed that they were interacting with an AI companion or a person, after the EEG experiment, we asked them to recall and report their belief about the identity of the joke-teller they had interacted with (Rheu et al., 2024). Sixty-three participants' responses aligned with the experimental condition (i.e., they identified their interlocutor as a human or an AI companion as intended), confirming the success of the manipulation; data from one participant in the human condition, who provided an irrelevant response ("a PC"), were excluded from data analyses.

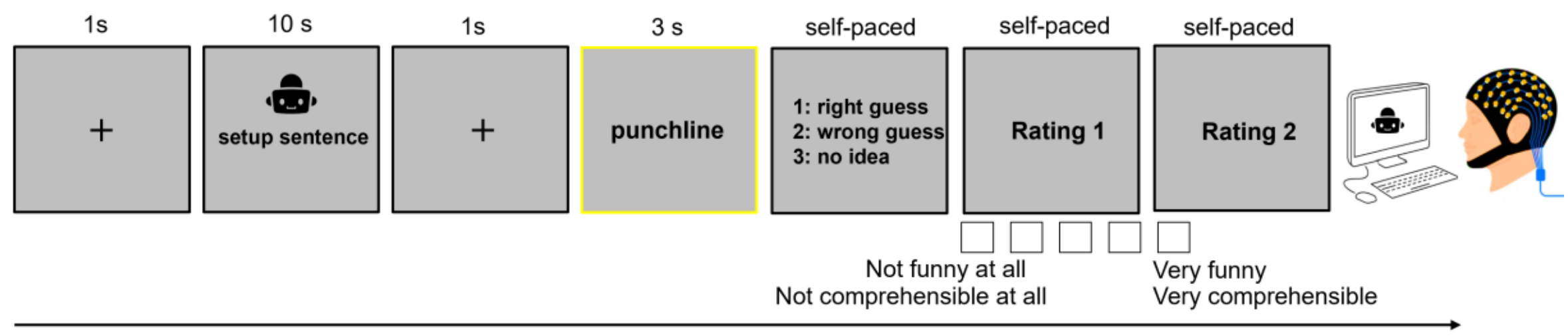

**Figure 1.** The paradigm of the EEG experiment.

### *2.4 EEG recording and preprocessing.*

EEG was collected from participants using 128 active Ag/AgCl electrodes, following an extended 10-20 system and online-referenced to the left earlobe. Signals were recorded with a g.HIamp amplifier and digitized at 512 Hz, with electrode impedances kept below 30 kΩ. Preprocessing involved customized scripts and the FieldTrip toolbox (Oostenveld et al., 2011) in MATLAB. Raw data were bandpass-filtered (0.1-30 Hz) and re-referenced to the average of both earlobes (Luck, 2014). For our high-density EEG data, independent component analysis was applied to the 1-30 Hz filtered data with dimensionality reduction to 30 components to identify and remove ocular artifacts (Luck, 2022; Ghandeharion & Erfanian, 2010; Winkler et al., 2011). Data were then epoched from 200 ms before to 1800 ms after critical word onset, with a baseline correction by subtracting the mean amplitude from 200 ms to 0 ms before critical word onset. Epochs exceeding ±150 μV were excluded (5.59% of trials). Following this procedure, six participants were excluded from subsequent data analyses because more than 30% of their trials contained excessive artifacts. Combined with the one participant who failed the manipulation check, the final sample comprised 57 participants (32 females, 25 males; mean age 22.6 years). In the primary analysis, trials where participants' guesses matched the punchline (31.26% of trials) were excluded, as the cognitive mechanisms for correctly guessing a punchline (likely involving predictive processing) are different from the incongruity-related process central to humor comprehension (Suls, 1972). Funniness ratings of the humorous stimuli also varied significantly between correct guesses ($M$ = 3.11, $SD$ = 1.25) and others ($M$ = 3.33, $SD$ = 1.32; $t$ = -3.52, $p$ < .001). Additionally, to evaluate the generalizability of our ERP findings, we conducted a follow-up robustness check by refitting the LME models on the full dataset, including trials where participants made incorrect guesses, correct guesses, or no guess at all.

## 3. RESULTS

### *3.1 Behaviorial results*

We first conducted linear-mixed effect (LME) modelling on funniness ratings, with Stimulus (control = -0.5, humor = 0.5) and Attribution (human = -0.5, AI = 0.5) as interacting fixed effects. Participant and Item were included as random effects (see Table S1 for model structures). For all LME analyses, we used the maximal random-effect structure justified by the data and determined by forward model comparison ($\alpha$ = 0.2, see Matuschek et al., 2017). The results showed a significant main effect of Stimulus ($\beta$ = 1.80, $SE$ = 0.11, $t$ = 16.29, $p$ < .001). As shown in Figure 2, humorous stimuli ($M$ = 3.33, $SD$ = 1.32) were rated as significantly funnier than control stimuli ($M$ = 1.54, $SD$ = 0.89). The interaction between Stimulus and Attribution was not significant ($\beta$ = 0.36, $SE$ = 0.20, $t$ = 1.74, $p$ = .088), indicating that the Stimulus effects (Humor minus Control) on funniness ratings were comparable across the human (1.63) and AI (1.94) conditions see Figure 2A.

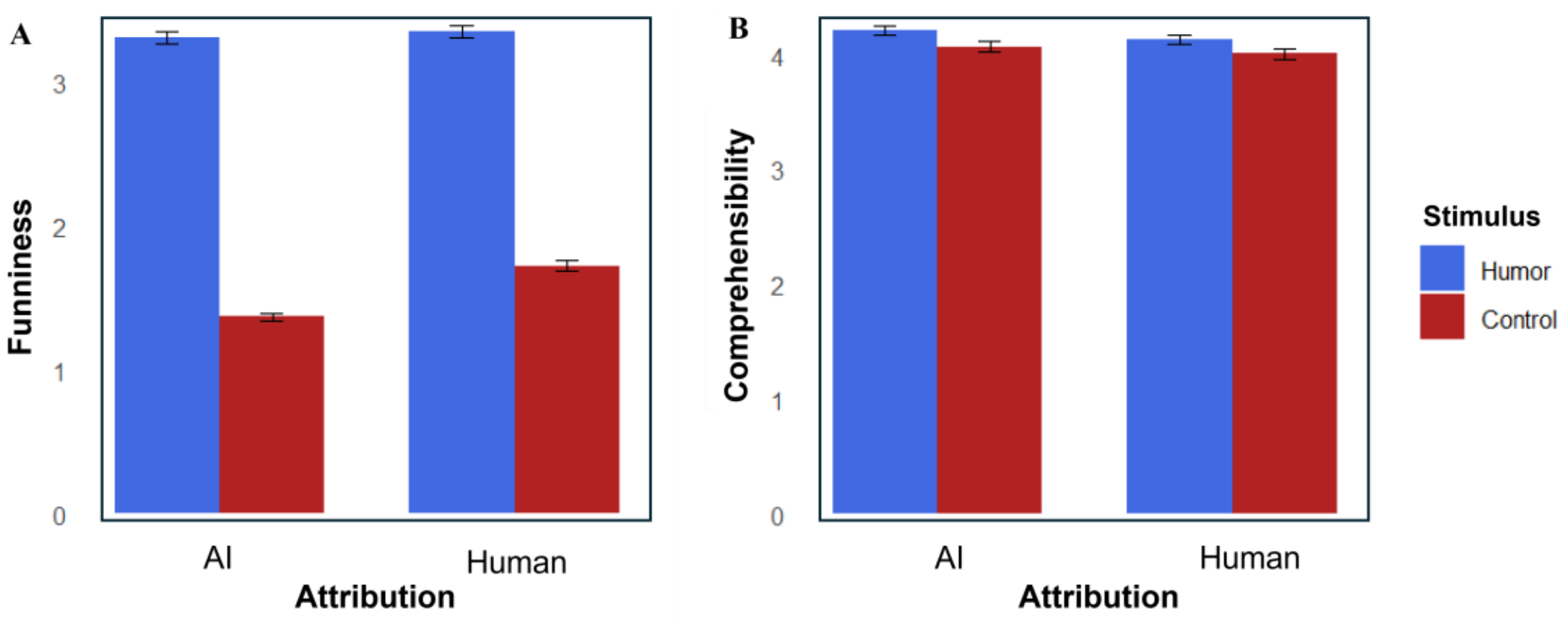

**Figure 2**. (A) Mean funniness ratings across Stimulus (humor vs control) and Attribution (human vs AI); (B) Mean comprehensibility ratings across Stimulus and Attribution. Error bars represent ±1 standard error of the mean (SEM).

We also carried out LME modelling on comprehensibility ratings, with Stimulus (control = -0.5, humor = 0.5) and Attribution (human = -0.5, AI = 0.5) as interacting fixed effects. Participant and Item served as random effects (see Table S1 for model structures). The results showed no significant main effect of Stimulus ($\beta$ = 0.08, *SE* = 0.07, *t* = 1.11, *p* = .272), indicating that humorous stimuli (*M* = 4.17, *SD* = 1.18) were rated as similarly comprehensible to control stimuli (*M* = 4.04, *SD* = 1.26). Additionally, the interaction between Stimulus and Attribution was not significant ($\beta$ = 0.04, *SE* = 0.11, *t* = 0.34, *p* = .735), suggesting that participants were able to comprehend humorous and control stimuli equally well, with no significant differences across the human (0.118) and AI (0.147) conditions (see Figure 2B).

### *3.2 Overview of EEG results*

Our analyses of ERP amplitudes focused on two key time windows after critical word onset: 300-900 ms and 1000-1800 ms. These windows were selected based on visual inspection of grand average ERP data (Figure 3) and established literature on the N400 (e.g., Kutas & Federmeier, 2011) and LPP (e.g., Dennis & Hajcak, 2009; Ku et al., 2017), respectively. The extended 300-900 ms window of the N400 captures the sustained negativity commonly observed in humor processing, reflecting prolonged cognitive effort during semantic conflict detection and resolution (Coulson & Kutas, 2001; Mayerhofer & Schacht, 2015). LME models were applied to the mean amplitudes within these windows for each trial, as LME methods are considered more robust than traditional ANOVA-based approaches in amplitude analyses (Frömer et al., 2018).

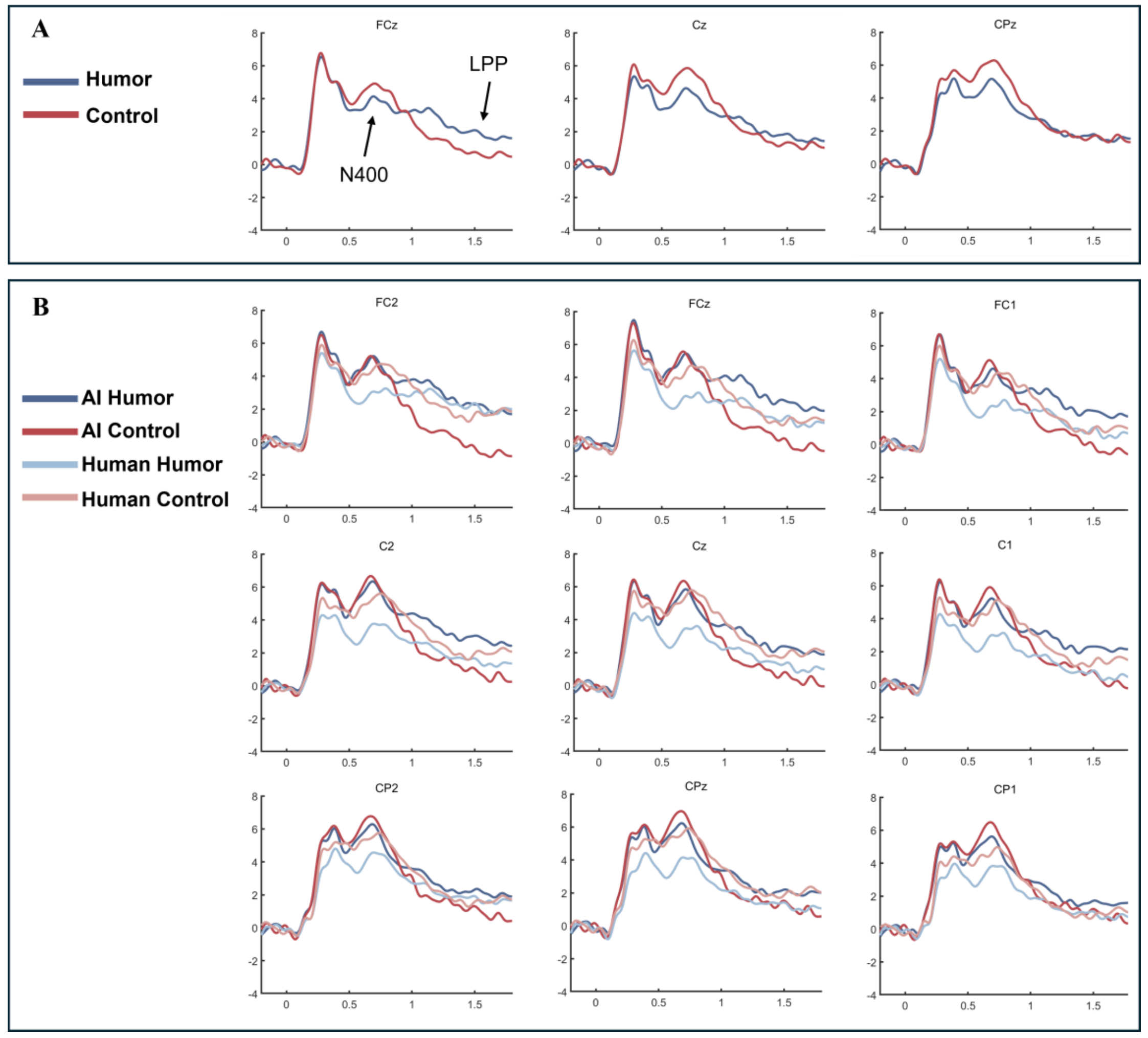

**Figure 3**. (A) Amplitudes of humors and controls; (B) Amplitudes of humors and controls in the human and AI conditions.

### *3.3 Topography analyses*

To investigate the topographical distribution of Stimulus (humor vs. control) effects on ERP amplitudes, we conducted posteriority and laterality analyses following earlier practice (e.g., Martin et al., 2016; Wu & Cai, 2026b). Scalp sites were divided into four regions: left-anterior, right-anterior, left-posterior, and right-posterior, with each region comprising 17 sites (see Table S2). Mean amplitudes were collapsed across all sites within each region.

For the posteriority analysis, we fit LME models with Stimulus, Attribution, and Posteriority (anterior = -0.5, posterior = 0.5) as interacting fixed effects (see Table S3 for model structures). In the 300-900 ms time window, neither the three-way interaction of Stimulus × Attribution × Posteriority ($\beta$ = -0.71, $SE$ = 0.59, $t$ = -1.21, $p$ = .225) nor the two-way interaction of Stimulus × Posteriority ($\beta$ = -0.05, $SE$ =0.29, $t$ = -0.19, $p$ = .853) reached significance. However, in the 1000-1800 ms window, a significant Stimulus × Posteriority interaction emerged ($\beta$ = -1.31, $SE$ = 0.40, $t$ = -3.30, $p$ < .001), indicating differential scalp topographies for the Stimulus effects (Humor minus Control). Subsequent analyses revealed a significant main effect of Stimulus ($\beta$ = 1.27, $SE$ = 0.51, $t$ = 2.47, $p$ = .016) and a significant Stimulus × Attribution interaction over anterior sites ($\beta$ = 2.61, $SE$ = 1.10, $t$ = 2.36, $p$ = .021). The interaction reflected a larger Stimulus effect for AI-attributed humor (2.48 μV) than for human-attributed humor (0.04 μV). Conversely, over posterior sites, neither the main effect of Stimulus ($\beta$ = -0.06, $SE$ = 0.55, $t$ = -0.12, $p$ = .907) nor the interaction between Stimulus and Attribution ($\beta$ = 1.68, $SE$ = 1.10, $t$ = 1.52, $p$ = .132) reached significance, suggesting comparable Stimulus effects across AI and human conditions.

Similarly, a laterality analysis was performed, using LME models with Stimulus, Attribution, and Laterality (right = -0.5, left = 0.5) as interacting fixed effects (see Table S3). In both the 300-900 ms and 1000-1800 ms windows, neither the three-way interaction of Stimulus × Attribution × Laterality (300-900 ms: $\beta$ = 0.09, $SE$ = 0.59, $t$ = 0.16, $p$ = .874; 1000-1800 ms: $\beta$ < -0.001, $SE$ = 0.79, $t$ = 0.00, $p$ > .999) nor the two-way interaction of Stimulus × Laterality (300-900 ms: $\beta$ = -0.09, $SE$ = 0.30, $t$ = -0.31, $p$ = .759; 1000-1800 ms: $\beta$ = -0.75, $SE$ = 0.40, $t$ = -1.88, $p$ = .060) reached significance.

Collectively, topographical analyses showed that the Stimulus effect was more pronounced over anterior sites than posterior sites in the 1000-1800 ms window, but not the 300-900 ms window, and exhibited no hemispheric asymmetry.

### *3.4 ROI analyses*

Based on the togography analyses, we selected two distinct ROIs (Region of Interest) for subsequent amplitude analyses. For the 300-900 ms window, the ROI comprised 49 whole-brain sites, while for the 1000-1800 ms window, the ROI consisted of 41 anterior sites (see Figure 4). In both cases, instead of collapsing the data within each ROI, we kept the individual channel data. This let us include Channel as a random effect in our LME models, consistent with the methods used by Ryskin et al. (2021) and Wu and Cai (2026b).

Our primary focus was to compare participants' neural responses to AI- versus human-attributed humor. We fit LME models with Stimulus and Attribution as interacting fixed effects for the mean amplitudes, and Participant, Item and Channel as random effects within each time window (see Table 2 for model structures). The results revealed a significant interaction between Stimulus and Attribution during 300-900 ms ($\beta$ = 1.24, *SE* = 0.10, *t* = 12.18, $p < .001$) and during 1000-1800 ms ($\beta$ = 2.51, *SE* = 0.17, *t* = 14.76, $p < .001$). In the 300-900 ms window, separate analyses showed a significant main effect of Stimulus in both the human ($\beta$ = -1.36, *SE* = 0.07, *t* = -18.39, $p < .001$) and AI conditions ($\beta$ = -0.22, *SE* = 0.08, *t* = -2.75, $p = .007$), such that humorous stimuli elicited more negative deflections compared to control stimuli (the N400 effect) in both conditions, while the N400 effect for AI-attributed humor (-0.26 μV) was smaller than that for human-attributed humor (-1.31 μV). In the 1000-1800 ms window, separate analyses showed significant main effects of Stimulus in both the human ($\beta$ = 0.25, *SE* = 0.12, *t* = 2.02, $p = 0.044$) and AI conditions ($\beta$ = 2.34, *SE* = 0.12, *t* = 19.05, $p < .001$), such that humorous stimuli elicited positive deflections compared to control stimuli (the LPP effect) in the later time window in both conditions, while the LPP effect for AI-attributed humor (2.68 μV) was significantly larger compared to that for human-attributed humor (0.27 μV).

To evaluate the generalizability of our ERP findings, we conducted a robustness check by refitting the LME models on the full dataset, including both incorrectly guessed and correctly guessed trials (see Table S4 for model structures). The results confirmed our primary findings, particularly regarding the distinct processing of AI-attributed humor. The results revealed a significant interaction between Stimulus and Attribution during 300-900 ms ($\beta$ = 1.28, *SE* = 0.52, *t* = 2.47, $p = .016$) and during 1000-1800 ms ($\beta$ = 2.18, *SE* = 0.14, *t* = 15.83, $p < .001$) after critical word onset. In the 300-900 ms window, separate analyses showed a significant main effect of Stimulus in both the human ($\beta$ = -1.53, *SE* = 0.06, *t* = -26.35, *p* < .001) and AI conditions ($\beta$ = -0.27, *SE* = 0.07, *t* = -4.09, *p* < .001), confirming that the N400 effect was present in both conditions but remained larger in the human condition. In the 1000-1800 ms window, separate analyses showed a main effect of Stimulus in the AI condition ($\beta$= 2.07, *SE* = 0.10, *t* = 20.75, *p* < .001), whereas the effect of Stimulus in the human condition was no longer significant ($\beta$ = -0.04, *SE* = 0.10, *t* = -0.44, *p* = .660). These results indicate that the unique neural signature of AI-attributed humor is robust across different levels of punchline predictability, whereas the late-stage response to human-attributed humor appears more contingent on the element of unpredictability (i.e., incorrect guesses).

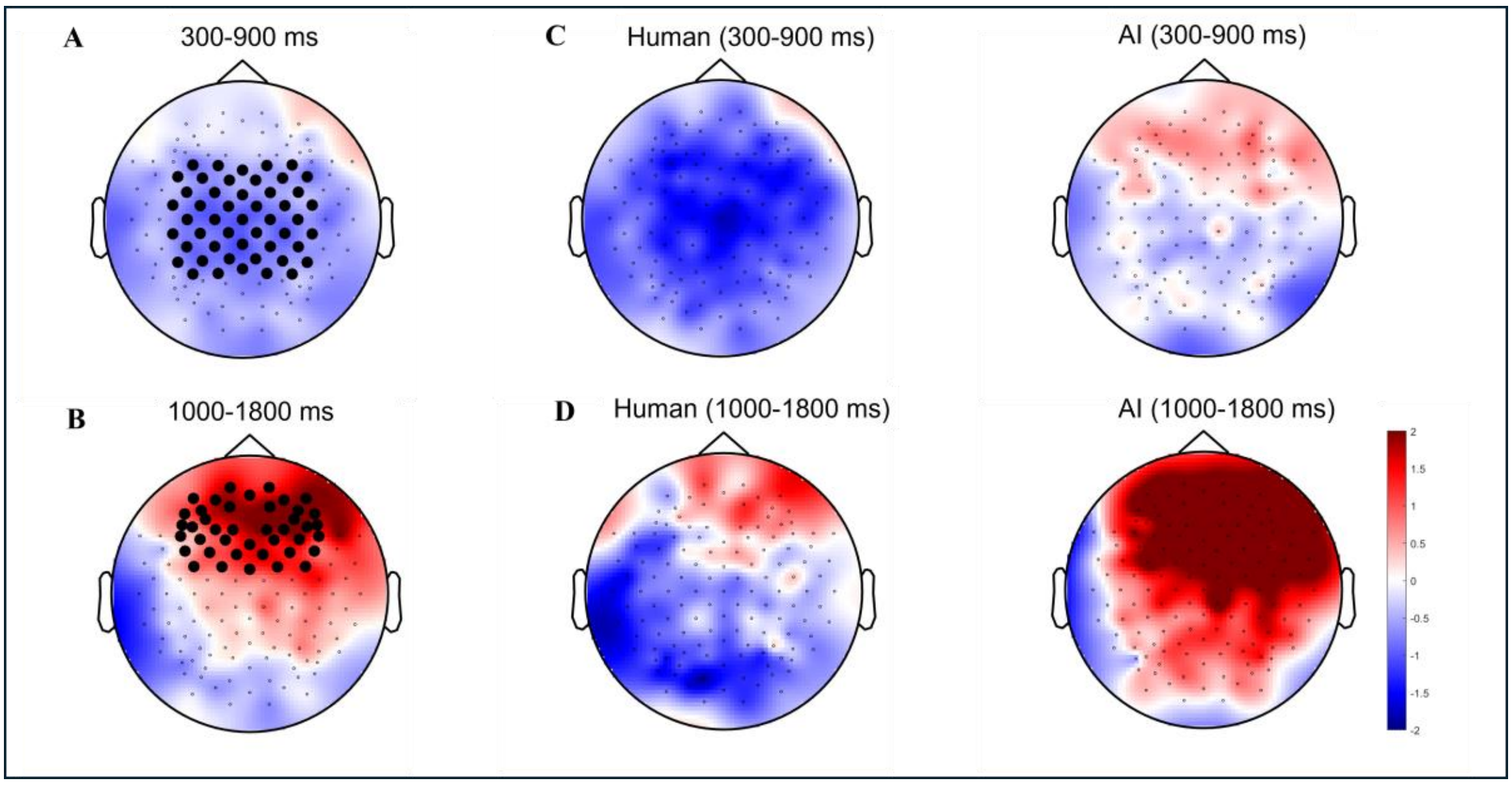

**Figure 4**. (A) The region used in the ROI analysis during 300-900 ms after critical word onset; (B) The region used in the ROI analysis during 1000-1800 ms after the critical stimulus onset; (C, D) Topographies of the Stimulus effects (Humor minus Control) during 300-900 ms and 1000-1800 ms after critical word onset. See Table S2 for a detailed list of channels.

**Table 2**. LME models for main ROI analyses.

| *Fixed effects* | *β* | *SE* | *t* | *p* |
|---|---|---|---|---|
| N400 (300-900 ms) | | | | |
| Intercept | 4.08 | 0.54 | 7.53 | < .001 |
| Stimulus | -0.74 | 0.05 | -13.54 | < .001 |
| Attribution | 0.70 | 1.00 | 0.70 | .485 |
| Stimulus : Attribution | 1.24 | 0.10 | 12.18 | < .001 |
| | | | | |
| LPP (1000-1800 ms) | | | | |
| Intercept | 1.12 | 0.55 | 2.03 | .046 |
| Stimulus | 1.30 | 0.09 | 14.92 | < .001 |
| Attribution | -0.62 | 1.01 | -0.61 | .542 |
| Stimulus : Attribution | 2.51 | 0.17 | 14.76 | < .001 |

Model for main amplitude analysis (300-900 ms): Amplitude ~ Stimulus * Attribution + (1 | Participant) + (1 | Item) + (Stimulus + 1 | Channel); Model for main amplitude analysis (1000-1800 ms): Amplitude ~ Stimulus * Attribution + (1 | Participant) + (1 | Item) + (1 | Channel)

As research shows that people dynamically adapt their mental model of the interlocutor (Wu et al., 2025), we further explored how ERP amplitudes varied over time, by fitting LME models with Stimulus, Attribution, and mean-centered Trial as interacting fixed effects (see Table S5 for model structures). The results showed a significant three-way interaction in both the 300-900 ms window ($\beta = 0.02$, $SE = 0.004$, $t = 5.79$, $p < .001$) and the 1000-1800 ms window ($\beta = 0.03$, $SE = 0.01$, $t = 4.09$, $p < .001$). Separate analyses within the 300-900 ms window showed that the Stimulus effect became significantly more negative over trials for both the AI condition ($\beta = -0.01$, $SE = 0.003$, $t = -3.79$, $p < .001$) and the human condition ($\beta = -0.03$, $SE = 0.003$, $t = -9.04$, $p < .001$). Notably, the rate at which the Stimulus effect became negative was more substantial in the human condition than in the AI condition. In the 1000-1800 ms window, the two conditions showed different patterns over time. For the AI condition, the Stimulus effect became significantly more positive as a function of trial ($\beta = 0.03$, $SE = 0.01$, $t = 5.70$, $p < .001$). In contrast, the Stimulus effect for the human condition remained stable, as the interaction between Stimulus and Trial was not significant ($\beta = -0.001$, $SE = 0.01$, $t = -0.32$, $p = .749$).

To investigate whether neural responses in the AI condition were modulated by participants' perceptions of AI trustworthiness and sincerity, we fit LME models with Stimulus and either Trustworthiness or Sincerity as interacting fixed effects (see Table S6 for model structures). Prior to analysis, the individual difference measures (Trustworthiness, Sincerity) were z-scored. In the 300-900 ms time window, as shown in Figure 5, we found that neural amplitudes in the AI condition significantly increased with perceived AI trustworthiness ($\beta$ = 2.05, *SE* = 0.76, *t* = 2.69, *p* = .012) and sincerity ($\beta$ = 1.43, *SE* = 0.66, *t* = 2.17, *p* = .040). Later, in the 1000-1800 ms time window, we observed a significant interaction between Stimulus and Trustworthiness ($\beta$ = 0.26, *SE* = 0.12, *t* = 2.06, *p* = .039). Specifically, the LPP effect for AI-attributed humor was significantly larger for participants with high AI trustworthiness (2.81 μV) compared to those with low AI trustworthiness (2.53 μV; see Figure 6). This interaction suggests that the LPP effect increases as a function of perceived AI trustworthiness. Neither the effect of Sincerity ($\beta$ = -0.07, *SE* = 0.77, *t* = -0.09, *p* = .933) nor the two-way interaction of Stimulus × Sincerity ($\beta$ = 0.28, *SE* = 0.61, *t* = 0.45, *p* = .656) in the 1000-1800 ms time window reached significance.

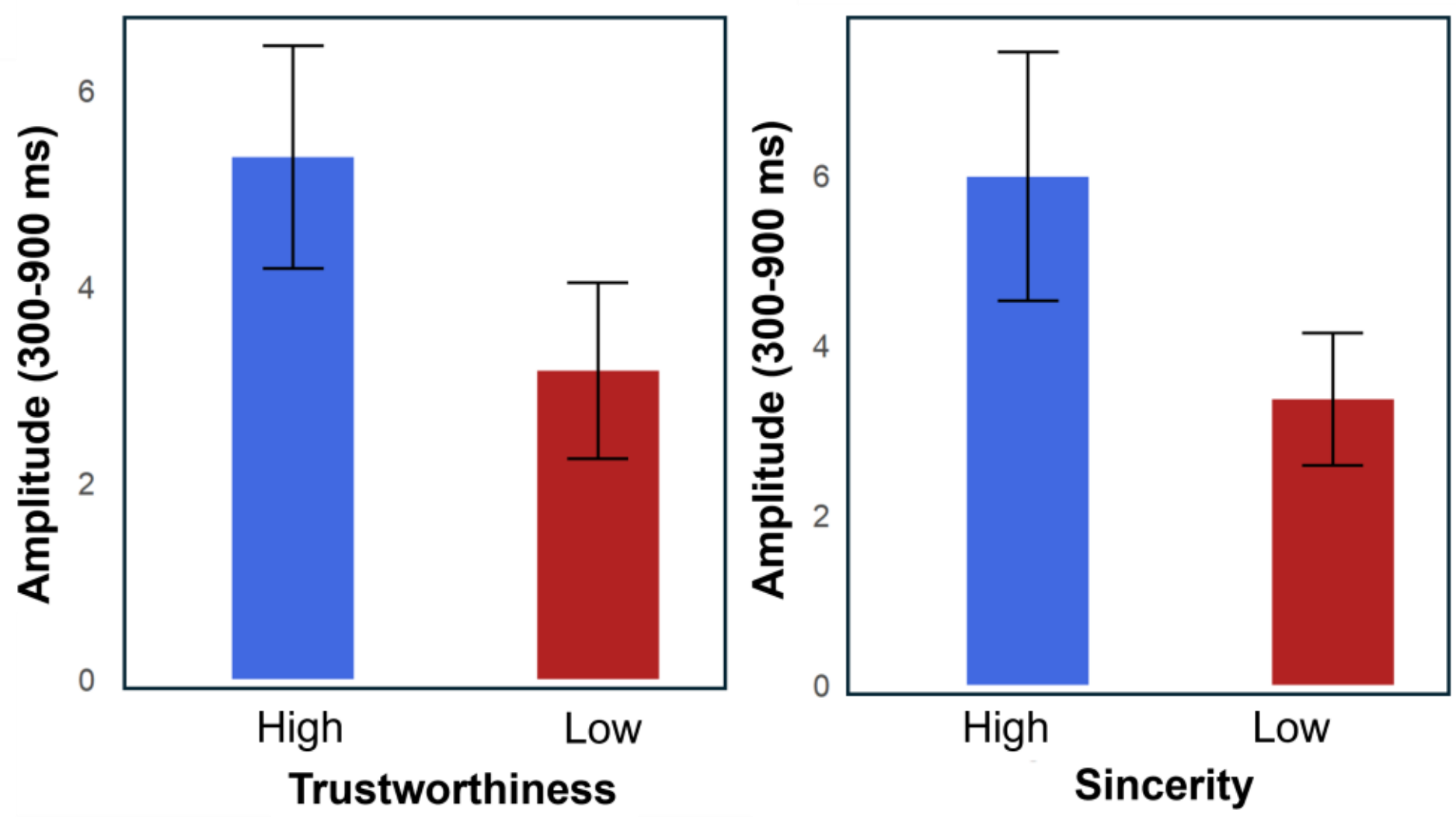

**Figure 5**. The modulative effect of Trustworthiness and Sincerity on ERP amplitudes in the 300-900 ms time window in the AI condition. Error bars represent ±1 standard error of the mean (SEM).

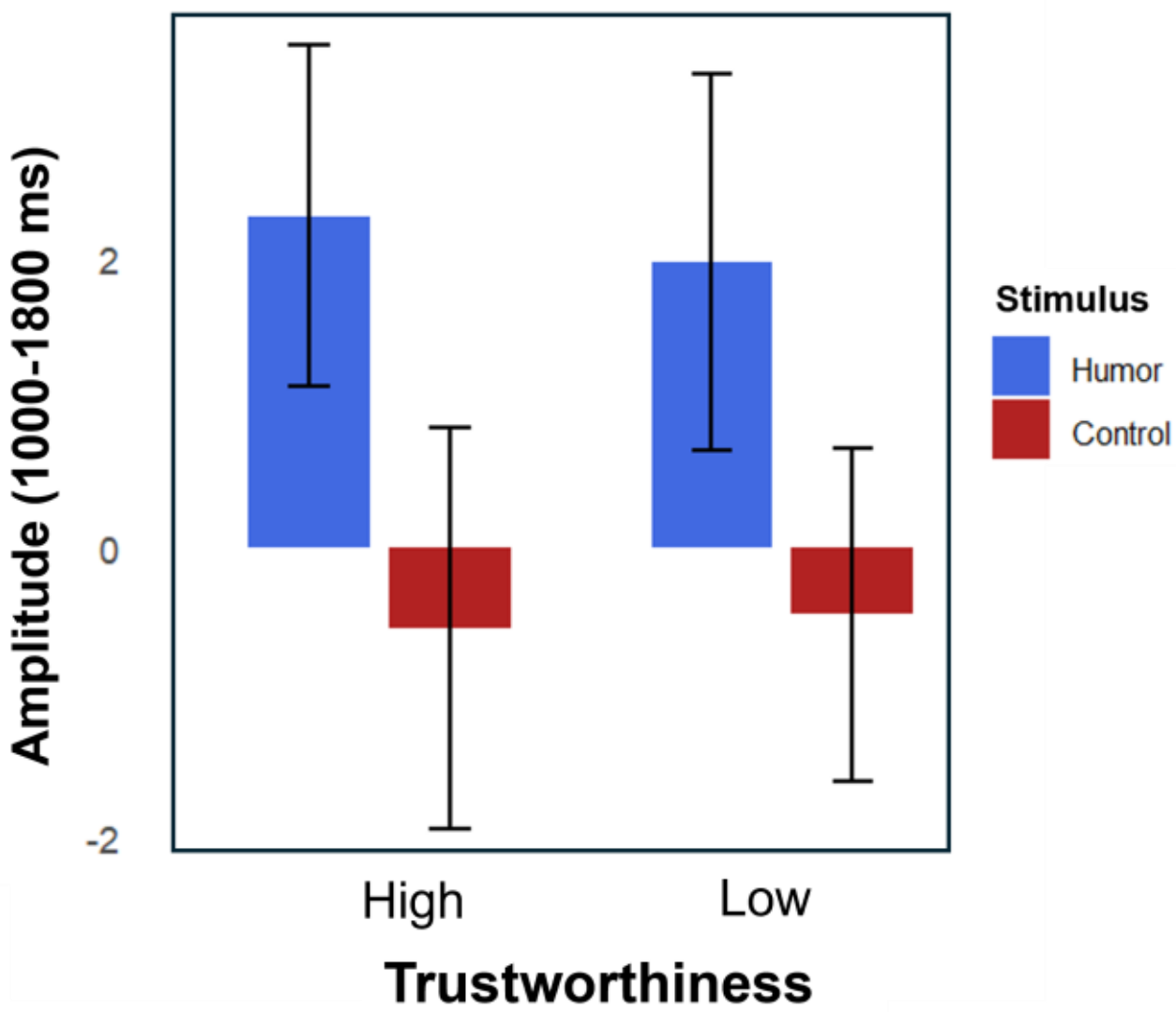

**Figure 6**. The modulative effect of Trustworthiness on ERP amplitudes across Stimulus in the 1000-1800 ms time window in the AI condition. Error bars represent ±1 standard error of the mean (SEM).

Moreover, to determine whether the positive deflections in the 1000-1800 ms time window were predicted by the perceived funniness of stimuli, we fit LME models with Attribution and z-scored Funniness as interacting fixed effects (see Table S7 for model structures). The results showed a significant main effect of Funniness in the 1000-1800 time window ($\beta = 0.40$, $SE = 0.03$, $t = 12.59$, $p < .001$), such that the positive deflections increased as a function of people's feeling of funniness (see Figure 7).

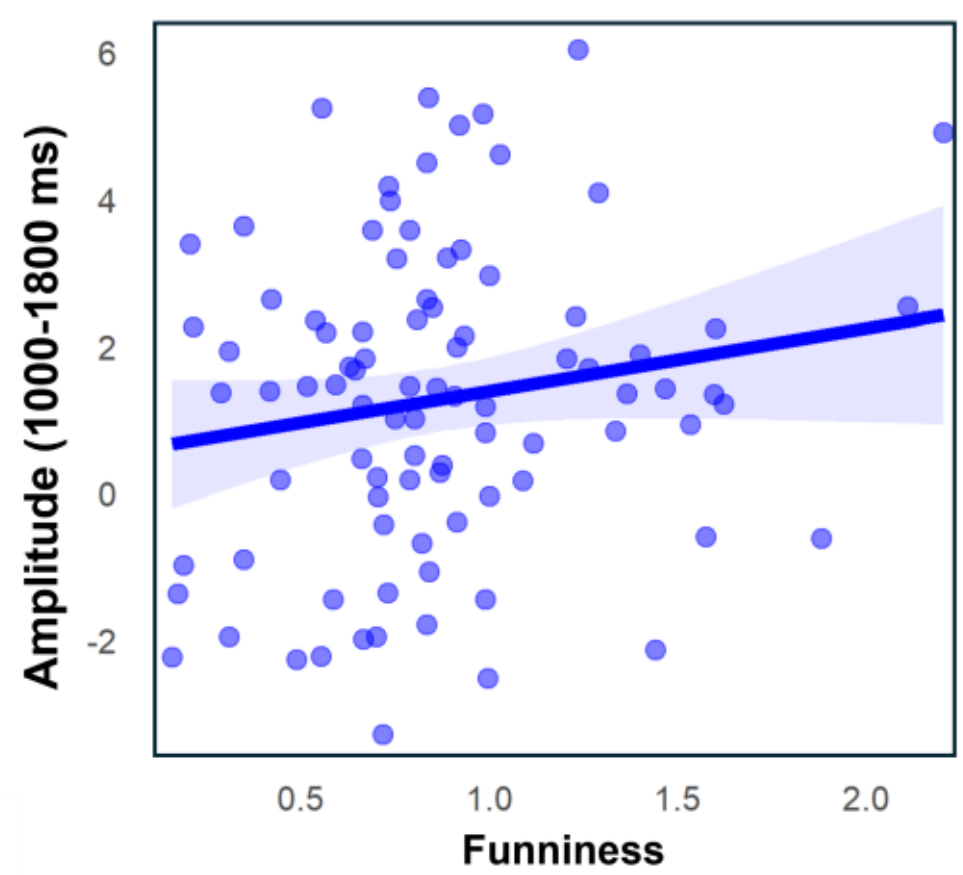

**Figure 7**. The modulative effect of Funniness on ERP amplitudes in the 1000-1800

ms time window.

### *3.5 Summary of EEG results*

Overall, AI- and human-attributed humorous stimuli elicited similar neural components of N400 and LPP. Nevertheless, the N400 effect for human-attributed humor was larger than that to AI-attributed humor, while the LPP effect for human-attributed humor was smaller than that to AI-attributed humor. As trials increased, the N400 for human-attributed humor became more negative, while the LPP remained stable. In contrast, the N400 for AI-attributed humor also became more negative, but to a lesser extent than the human condition, while the LPP became more positive over time. Furthermore, higher perceived AI trustworthiness and sincerity predicted reduced N400 amplitudes across all stimuli, whereas late-stage LPP for AI-attributed humor increased as a function of perceived AI trustworthiness. Additionally, higher funniness ratings predicted increased LPP across all stimuli in the 1000-1800 ms time window.

## 4. DISCUSSION

The present study investigated people's neural responses when comprehending AI- versus human-attributed humor. Behavioral analysis of funniness ratings revealed that humorous stimuli were rated as funnier than control stimuli, validating the effectiveness of our humor manipulation. Notably, there was no significant difference in funniness ratings between AI- and human-attributed humor, indicating that on a conscious level, participants found the humor comparably funny regardless of the attributed source, consistent with previous findings (e.g., Bower & Steyvers, 2021). One possibility is that algorithm aversion is readily countered by direct evidence of the agent's competence (Bower & Steyvers, 2021). Furthermore, this effect may be amplified by a violation of low expectations regarding AI's linguistic and social competence; the delivery of culturally embedded puns likely generates a novelty effect that increases perceived funniness, offsetting potential biases against artificial sources. We observed that the same ERP components were present in both human-AI and human-human humorous interactions, indicating that the brain employs comparable neural mechanisms for semantic conflict processing and affective and evaluative processing regardless of the perceived source. This similarity broadly aligns with recent research suggesting that AI can engage in social and emotional interaction as viable social companions (Inzlicht et al., 2024; Ovsyannikova et al., 2025; Rubin et al., 2025). Despite these similarities, the magnitude and temporal dynamics of these components differed between conditions, suggesting that the perceived source modulates the underlying neurocognitive depth of these processes.

Our results showed that AI-attributed humor elicited a smaller N400 effect than human-attributed humor. As the N400 effect is suggested to reflect cognitive effort during semantic conflict processing in humor comprehension (Coulson & Kutas, 2001; Du et al., 2013; Ku et al., 2017; Mayerhofer & Schacht, 2015), the reduced N400 indicates that participants processed AI-attributed humor with lower cognitive demand. The observed sustained N400 likely encompasses both semantic conflict detection and resolution phases (Coulson & Kutas, 2001; Mayerhofer & Schacht, 2015). The reduced N400 for AI-attributed humor indicates a diminished allocation of cognitive effort toward semantic conflict detection. This reduced effort may stem from two sources of expectations. First, participants may have approached AI-attributed humor with reduced expectations for semantic coherence and logical consistency, consistent with previous findings on AI-attributed semantic anomalies (Rao et al., 2025b). Second, participants may have held lower initial expectations for an AI's capacity to deliver culturally embedded humor. This reduction in perceived interlocutor competence likely led to decreased resource allocation during early-stage semantic conflict detection. This initial reduced effort likely extends into the resolution phase; the attenuated sustained N400 may suggest a tendency toward shallow resolution, wherein participants may be less motivated to resolve semantic conflict and infer pragmatic meaning when the humor is attributed to an AI agent.

Complementing this view, the interpretation of these ERP components should also account for the "prediction-first" nature of our experimental paradigm. Because participants spent up to 10 seconds actively anticipating the punchline, the N400 likely reflects neurophysiological responses to feedback, such as prediction error and hypothesis updating (Kuperberg & Jaeger, 2016; Rabovsky et al., 2018). In this context, the reduced N400 for AI-attributed humor may reflect diminished precision-weighting of prediction errors. Predictive coding accounts posit that the brain weights the importance of prediction errors based on the perceived reliability/competence or precision of the information source (Kuperberg & Jaeger, 2016). Because participants likely perceived AI as less competent than humans in social interaction, the brain assigned lower weight to the prediction errors triggered by AI's feedback, leading to an attenuated N400. Similarly, the attenuated N400 may indicate reduced hypothesis updating; specifically, they may have been less inclined to revise their internal representations when their initial guesses proved incorrect. Taken together, these results suggest that the perceived source may modulate how the brain processes feedback when active predictions fail to materialize. These two perspectives (i.e., semantic conflict processing and predictive feedback) are likely interrelated. It is possible that reduced conflict detection and attenuated feedback-related processing observed here may be the very mechanisms that lead to a shallower resolution of AI-attributed humor.

Conversely, our findings revealed that AI-attributed humor elicited a larger LPP effect than human-attributed humor. We observed that perceived funniness positively predicted LPP amplitudes in both the AI and human conditions; this provides evidence that the late-stage positivity tracks the subjective humor experience. The LPP is often discussed as a marker of the affective response to humor (Chang et al., 2018), but it is also sensitive to non-emotional factors including novelty and surprise (Ferrari et al., 2016), attentional allocation (Ferrari et al., 2008;), social expectancy violations (Bartholow et al., 2001), and source-related social evaluation (Schindler et al., 2019). Therefore, the enhanced LPP observed here likely reflects intensified late-stage affective and evaluative processing during humor appreciation in the AI condition. From an affective perspective, these results suggest that AI-attributed humor amplifies affective neural engagement, adding a physiological dimension to the study of algorithm aversion (Castelo et al., 2019; Mahmud et al., 2022). While our behavioral findings showed comparable funniness ratings, the larger LPP for AI-attributed humor may suggest that the brain encodes a distinction between perceived interlocutors, which is not reflected in behavioral reports. From an evaluative perspective, in our "prediction-first" paradigm, these non-emotional factors likely converge to index the updating of hypotheses regarding the perceived source's capabilities. Specifically, because an AI agent that can deliver culturally embedded jokes may be perceived as novel or expectancy-violating, this triggers a more intensified re-evaluation of the interlocutor model as the brain reconciles AI's mastery of complex cultural constraints with its prior expectations.

The dynamic changes observed over the course of the experiment revealed tentative insights into how the brain adapts to different perceived social agents over time. Our exploratory trial-level analyses revealed two divergent neurocognitive trajectories. In the human-human humorous interaction, while the LPP effect remained stable, the N400 effect became more pronounced over trials. This suggests that as the experiment progressed, participants may have required greater cognitive effort in semantic conflict processing when encountering human-attributed humor, possibly reflecting a shift in processing efficiency or a change in engagement with a familiar interlocutor. Alternatively, the more pronounced N400 might indicate intensified feedback-related processing. As the interlocutor model of the human partner stabilized, the brain may have assigned higher precision-weighting to the human joke-teller, potentially intensifying the neural responses to prediction errors. In contrast, in response to AI-attributed jokes, neural responses followed a different trajectory. Here, we observed that although the trajectory of human-AI interaction also trended toward a more negative N400, this change occurred at a smaller rate than during human-human interaction. This may suggest that the cognitive effort allocated to semantic conflict processing increases less steeply for AI- than human-attributed humor, or that the precision-weighting of the AI joke-teller does not intensify as rapidly as it does for the human joke-teller. Furthermore, AI-attributed humor elicited progressively enhanced

LPP effects, possibly indicating intensified late-stage affective and evaluative processing as the experiment progressed. It is likely that because our stimuli relied on culturally embedded puns, each successful AI-attributed joke delivery acted as a salient violation of prior low expectations. Cumulative expectancy violations from these unexpected successes may progressively increase the magnitude of the mental model update. Furthermore, it is possible that this repeated violation of low expectations may amplify the affective response over time. Speculatively, this neurophysiological evidence provides a potential mechanistic explanation for previous behavioral findings indicating that people's appreciation for AI-generated humor increases with repeated exposure (Joshi, 2025). However, these findings are preliminary; further research is required to determine whether these dynamic shifts consistently characterize human-AI social interaction.

This dynamic adaptation to AI-attributed humor provides a potential mechanism for mitigating algorithm aversion. While the literature suggests a critical "caring" deficit in human-AI interaction (Rubin et al., 2025), our findings support the idea that humor is a domain where AI's competence can neutralize initial biases (Bower & Steyvers, 2021). Specifically, participants who viewed AI more trustworthy showed a larger LPP, suggesting that favorable social perceptions toward AI may intensify late-stage affective and evaluative processing with the attributed agent. For these individuals, AI's successful delivery of culturally embedded puns likely triggers a more intensified update of the interlocutor model. This may reflect a process where participants who perceived AI as more trustworthy are more open to recalibrating their perceptions of the agent, allowing AI's success to challenge the "robotic" stereotype and expand the agent's perceived social agency. This phenomenon extends to the processing of AI language more broadly. Our results showed that the N400 effect in response to both AI-attributed humor and control decreased as a funciton of perceived AI trustworthiness and sincerity, indicating decreased cognitive effort during semantic integration. This is likely because the participants' positive social perceptions about an AI agent may have lowered the threshold for semantic integration. With a mental model of an AI agent as a sincere and trustworthy interlocutor, the brain may have been better primed to integrate in its language, reducing the cognitive effort required to retrieve and integrate semantic meaning. This pattern indicates that social perceptions about the interlocutor, independent of the communicative content itself, modulate early-stage semantic conflict processing and late-stage affective and evaluative processing. This provides concrete evidence for interlocutor modelling effects (Cai et al., 2017; Wu & Cai, 2026; Wu et al., 2024), demonstrating that people's mental models of their conversation partners, even non-human ones, influence humor comprehension at the neural level.

Our findings also align with a broader psychological phenomenon whereby people spontaneously attribute human-like characteristics to artificial agents. Previous research has documented this anthropomorphic tendency across multiple dimensions, showing that people attribute traits such as age (Dou et al., 2021; Powers & Kiesler, 2006), gender (Powers et al., 2005; Powers & Kiesler, 2006), language background (Cowan et al., 2019), language competence (Dunn & Cai, 2025; Shen & Wang, 2023), and perspective taking capabilities (Loy & Demberg, 2023) to AI systems and adjust their interaction accordingly. These attributions suggest that people conceptualize AI as possessing specific knowledge bases and social profiles, which in turn shapes their expectations and processing of AI-generated content.

These findings may be particularly salient for culturally embedded humor. As our stimuli often relied on socially shared idiomatic knowledge or online community slang, the observed neural patterns for AI may be driven by the expectancy-violating nature of an AI delivering such humor. It remains to be seen whether superficial homophonic puns, which require phonological shifts rather than deep cultural retrieval, would elicit a similar magnitude of interlocutor model updating. Also, the "prediction-first" nature of our active guessing task likely recruited distinct cognitive processes compared to passive humor comprehension. Future studies should employ paradigms with varying predictive demands to further disentangle the neurophysiological responses to task-related feedback from those associated with humor processing.

These findings have important implications for human-AI interaction design. Our neural evidence demonstrates that humorous interactions with an AI agent elicit more intensified late-stage affective and evaluative processing than human-human humorous interactions, with this effect increasing with repeated exposure. This suggests humor represents a promising domain for overcoming algorithm aversion, as users exhibit genuine neural engagement with AI despite potential explicit biases. Furthermore, the finding that positive social perceptions of AI facilitate semantic processing offers specific design guidance: AI systems that cultivate positive social perceptions may benefit from enhanced processing efficiency and deeper user engagement by leveraging these top-down expectations. Nevertheless, the implementation of AI joke-tellers must also recognize when authentic human connection is paramount. Human-generated content may be more beneficial in some social situations requiring genuine emotional resonance, shared vulnerability, or personal understanding (Rubin et al., 2025). The challenge lies in identifying these critical moments where human humorous expression carries irreplaceable social and emotional value, allowing AI humor to complement rather than replace meaningful human interaction. This mechanistic understanding facilitates more targeted approaches to developing socially competent AI companions that enhance, rather than supplant, human social bonds.

## 5. CONCLUSION

While humor elicits similar core neural responses for early-stage semantic conflict processing and late-stage affective and evaluative processing regardless of whether it is attributed to an AI agent or a human being, the magnitude and dynamic patterns of these responses reveal that AI-attributed humor evoked a sustained smaller N400 compared to human-attributed humor, suggesting reduced cognitive effort in conflict detection and resolution, or attenuated feedback-related processing. Moreover, AI-attributed humor evoked a larger LPP, indicating an intensified late-stage affective and evaluative processing. This neural-level sensitivity, which appears to bypass the biases of algorithm aversion, likely reflects the brain's dynamic adaptation and the enhanced saliency of humor from a novel source. This underscores that a person's mental model of their conversation partner, even a non-human one such as an AI agent, actively influences humor comprehension at a neural level. Ultimately, our findings suggest that humor is a potent tool for fostering genuine user engagement and serves as a promising domain for mitigating initial algorithmic biases in human-AI social interactions.